\documentclass{article}
\usepackage{graphicx}%
\usepackage{multirow}%
\usepackage{amsmath,amssymb,amsfonts}%
\usepackage{amsthm}%
\usepackage{mathrsfs}%
\usepackage[title]{appendix}%
\usepackage{xcolor}%
\usepackage{textcomp}%
\usepackage{manyfoot}%
\usepackage{booktabs}%
\usepackage{algorithm}%
\usepackage{algorithmicx}%
\usepackage{algpseudocode}%
\usepackage{listings}%
\usepackage{caption}
\usepackage{subcaption}
\usepackage{bookmark}

\usepackage[utf8]{inputenc}
\usepackage{natbib}
\usepackage{booktabs}
\setcitestyle{authoryear,open={(},close={)}}
\usepackage{csquotes}
\usepackage{hyperref}
\usepackage{booktabs}
\usepackage{makecell}

\DeclareMathOperator*{\argmax}{argmax}
\DeclareMathOperator*{\myopicvoc}{MYOPIC\_VOC}

\title{An intelligent tutor for planning in large partially observable environments}

\author{Lovis Heindrich$^1$$^*$, Saksham Consul$^1$, Falk Lieder$^{1, 2}$}
\date{
    $^1$Max Planck Institute for Intelligent Systems, Tübingen, Germany\\
    $^2$Department of Psychology, UCLA, Los Angeles, USA\\
    $^*$E-mail: lovis.heindrich@tuebingen.mpg.de\\[2ex]
    \today}

\begin{document}

\maketitle

\abstract{ 
AI can not only outperform people in many planning tasks, but it can also teach them how to plan better. A recent and promising approach to improving human decision-making is to create intelligent tutors that utilize AI to discover and teach optimal planning strategies automatically. Prior work has shown that this approach can improve planning in artificial, fully observable planning tasks. Unlike these artificial tasks, many of the real-world situations in which people have to make plans include features that are only partially observable. 
To bridge this gap, we develop and evaluate the first intelligent tutor for planning in partially observable environments. Compared to previous intelligent tutors for teaching planning strategies, this novel intelligent tutor combines two innovations: 1) a new metareasoning algorithm for discovering optimal planning strategies for large, partially observable environments, and 2) scaffolding the learning process by having the learner choose from an increasing larger set of planning operations in increasingly larger planning problems. We found that our new strategy discovery algorithm is superior to the state-of-the-art. A preregistered experiment with 330 participants demonstrated that the new intelligent tutor is highly effective at improving people's ability to make good decisions in partially observable environments. This suggests our intelligent cognitive tutor can successfully boost human planning in complex, partially observable sequential decision problems. That makes the work presented in this a promising step towards using AI-powered intelligent tutors to improve human planning in the real world.
}



\maketitle
\newpage

\section{Introduction} 

Intelligent tutoring systems (ITS) have long been a promising approach to facilitating student learning by combining deliberate practice with personalized feedback according to cognitive models of the student's learning progress \citep{corbett1997intelligent, graesser2012intelligent}. Compared to human tutors, ITS have been shown to be similarly effective \citep{vanlehn2011relative, anderson1985intelligent}, and offer an efficient way to scale personalized tutoring to large numbers of learners \citep{koedinger1997intelligent}. While intelligent tutoring systems have in the past mainly been applied to teach concrete problem-solving skills in subjects like computer science or mathematics \citep{mousavinasab2021intelligent}, more recent work has proposed using intelligent tutors to improve more general (meta)cognitive skills such as planning \citep{aleven2006toward, guerra2017book, roll2005modeling, Callaway2021Leveraging, chi2010meta}.


Planning is crucial for many important real-world decisions that require foresight. Such decisions are critical to people's success in many domains of life, including education, work, health, finances, and family. In all of these domains, there is significant room for improvement because people have been shown to make many systematic errors that could be avoided by relying on more far-sighted decision strategies. 
It might be possible to help people avoid such mistakes by teaching them better decision-making strategies \citep{larrick1990teaching, gigerenzer2011heuristic, hertwig2017nudging, hafenbradl2016applied}. 
Recent research suggested that people's planning strategies make near-optimal use of their cognitive resources in some situations, but are systematically suboptimal in others \citep{Callaway2022Rational,Jain2022}. Building on these findings, subsequent research showed that human planning can be significantly improved by using intelligent tutors to teach people optimal planning strategies \citep{Callaway2021Leveraging,consul2021improving,mehta2022leveraging}. 

In contrast to outsourcing people's decisions to computers, this approach enables people to effectively plan in novel future situations where they have to think for themselves \citep{Callaway2021Leveraging,consul2021improving,mehta2022leveraging}. 
In this line of work, the optimal planning strategies were discovered by leveraging artificial intelligence. Concretely,
using a model of the planning task and the elementary operations of human planning, methods from reinforcement learning were fashioned into algorithms for computing near-optimal cognitive strategies for human planning \citep{callaway2017learning, mehta2022leveraging, consul2021improving}. These algorithms have been extended to increasingly larger planning problems. 
Being able to algorithmically discover planning strategies that perform better than people's intuitive planning strategies opens up the possibility of teaching these strategies to people. This idea has been successfully applied in previous work, where human planning was improved in a fully automated way through training sessions with an intelligent tutoring system \citep{Callaway2021Leveraging, consul2021improving} that taught people the metacognitive awareness of what information they already possess and which information they need to acquire. 

While intelligent tutoring systems that utilize strategy discovery methods are currently limited to relatively small and simplified planning problems, a major next step in this line of research is to extend strategy discovery methods to real-life scenarios. Solving decision tasks closer to real-life problems requires scaling current methods to more complex and naturalistic decision problems. Being able to model and discover efficient planning strategies for such problems has the large potential of having a positive impact on people's lives by improving their planning in relevant issues people face in the real world, such as deciding what to work on next or which charity to donate to \citep{caviola2021psychology}.

This article advances computer-based learning systems by introducing general principles for the development of cognitive tutors that leverage AI to discover, represent, and teach cognitive strategies in environments featuring uncertainty. 
Our general framework fits many real-world cognitive tasks central to the learning objectives of schools and universities, such as time management, choosing what to work on, mathematical problem-solving, and general planning. Our framework extends the theoretical foundations of intelligent tutoring systems by applying a refined, computational version of rational analysis \citep{anderson1990adaptive} to derive which cognitive strategies the intelligent tutoring system should teach \citep{Callaway2021Leveraging}. This overcomes the key bottleneck of designing intelligent tutors for cognitive skills: identifying effective cognitive strategies and breaking them down into a precise sequence of steps students can be taught to follow, a process that usually relies on manually designed cognitive strategies \citep[e.g.][]{aleven2006toward, chi2010meta, ozsoy2009effect}. This is especially important as there are many cognitive skills for which no human instructor can describe the optimal strategy with sufficient precision for use in a cognitive tutor.

Specifically, we extend intelligent tutoring systems for human planning \citep{Callaway2021Leveraging}
to partially observable environments. It thereby overcomes a crucial limitation of previous research on automatically improving human planning, namely, the fact that it was conducted in simplistic planning tasks where the environment is fully observable. In the real world, making good decisions often requires farsighted planning under uncertainty. Thinking about a future action or event is unlikely to lead to a completely reliable prediction of its consequences. For example, when planning which bus to take, 
the precise time of arrival cannot be guaranteed since the bus can be delayed by a wide range of random events, such as traffic jams. One, therefore, needs to make, refine, and plan with uncertain estimates of the bus timings. The efficiency of people's metacognitive strategies in such more realistic scenarios remains unknown. Moreover, while computer algorithms for planning in partially observable environments have been studied for a long time \citep{monahan1982state, sondik1971optimal, kaelbling1998planning}, there is no prior work on discovering metacognitive strategies for human planning in partially observable environments \citep{Callaway2021Leveraging,consul2021improving,mehta2022leveraging}. 

This article reports the development and evaluation of our new intelligent tutor for teaching people how to plan under uncertainty.  We first formalize the skill taught by our tutor with a computational model of planning in partially observable environments. We then introduce the first algorithm that can discover efficient cognitive strategies for human planning in partially observable metareasoning environments. 
Third, we develop an adaptive and intelligent tutoring system that teaches people to use the discovered strategies automatically using feedback. Fourth, we demonstrate that the general methods developed in this article can be used to effectively teach people cognitive strategies in partially observable environments by applying our framework to discover and teach cognitive strategies in a large online experiment. Our results show that the planning strategies people intuitively use in partially observable environments are highly suboptimal, and that teaching people the planning strategies discovered by our method substantially improves the quality of their decisions. This constitutes an important step towards improving human decision-making in the real world. 

Our general approach to teaching automatically discovered cognitive strategies through practice with feedback is highly relevant to the design of intelligent tutoring systems for a wide range of cognitive skills taught in schools and universities around the world. There are many cognitive skills students commonly learn through practice with feedback, for example, solving word problems in math, physics, and chemistry, programming, algebra, and logical reasoning. There are many intelligent tutoring systems that support this process by choosing appropriate practice problems and providing students with feedback, such as Carnegie Learning’s MATHia \citep{ritter2007cognitive}, LISP Tutor \citep{anderson1985lisp}, and AutoTutor \citep{graesser2012autotutor}. In the future, we hope that the principles and computational methods introduced and evaluated in this article can be used to improve these real-world applications of helping students learn cognitive skills through practice with feedback by discovering efficient cognitive strategies automatically and utilizing the computational model of optimal metareasoning to provide high-quality feedback.


The remainder of this article is structured as follows. Section~2 introduces the technical background of our approach to developing intelligent tutors: metareasoning and meta-level MDPs. Section~3 formalizes the skill to be taught, namely planning under uncertainty, and presents our new strategy discovery algorithm. Section~4 benchmarks our new strategy discovery algorithm against two baseline algorithms. Section~5 presents our intelligent cognitive tutor that teaches the discovered strategies to humans and describes our human training experiment in which the tutor successfully improved people's planning strategies. Lastly, Section~6 discusses our work's limitations and future implications.

\section{Background}

When planning, human decision-makers are inherently limited by their cognitive resources and their limited time. Good planning strategies make efficient use of these limited cognitive resources to arrive at good plans with a limited amount of computation. This idea has been formalized by the theory of \textit{resource-rationality} \citep{LiederGriffiths2020}. The resource-rationality of a planning strategy is the average expected return of the resulting plan minus the average cost of the time and computational resources spent on computing the plan. This formal measure can be used 
as a metric for measuring the quality of human planning and comparing it against automatically discovered planning strategies \citep{consul2021improving}.
The human planning process can formally be described as a meta-level Markov Decision Process \citep{hay2014selecting, Callaway2022Rational}, which is an extension of Markov Decision Processes (MDP). MDPs are a general framework used to simulate and evaluate decision-making and described by a tuple $(\mathcal{S}, \mathcal{A}, P, r)$, in which an agent interacts with a (stochastic) environment in state $s\in\mathcal{S}$ by performing actions $a\in\mathcal{A}$ and, in return, receiving a reward $r(s,a)$ and a new environment state $s'$ sampled from the transition function $P(s, a)$. 

Utilizing the framework of the meta-level MDP allows separating a decision problem into an object-level component, which models the interaction with the real world, and a meta-level component, which models the (often internal) planning operations used to decide on object-level actions \citep{hay2014selecting}. Based on the tradition in Artificial Intelligence to distinguish between object and meta-level reasoning \citep{RUSSELL1991361}, this split of a decision problem accurately captures the qualitative difference between the external world and internal beliefs. Modeling the two problems separately allows defining the metareasoning process in a clear and legible way that can be studied in isolation.

In a meta-level MDP, the state represents the agent's current belief about the environment (\textit{belief state}). Actions in the meta-level MDP are \textit{computations}, which reveal information about the environment while incurring a planning cost. An additional meta-level action, the termination action, represents stopping the planning process and acting out the plan with the highest expected return. Achieving a high resource-rationality score (\textit{RR-score}) requires efficiently balancing between gathering information and deciding when to stop planning.

In an educational setting, an example of a meta-level planning problem is choosing an approach to solve a physics problem. Computations refer to the mental operations involved in deciding how to solve the given problem, such as evaluating the effectiveness of different solution methods, assessing whether the problem can be broken down into sub-problems, or thinking about which material might be helpful for solving the problem. Object-level actions, on the other hand, refer to the concrete actions used to solve the problem, such as calculating intermediate variables or applying a formula. The belief state represents the student's internal understanding of the problem, as well as their confidence in different solution approaches. Resource-rational metareasoning is important in this context, as directly attempting to solve the problem with the first method that comes to mind could prove inefficient or impossible, but spending excessive amounts of time weighing different options wastes mental resources and time. Prior work \citep{ozsoy2009effect} has shown that training general and hand-crafted metacognitive skills such as asking students to evaluate their understanding of the problem at hand can significantly improve students' test scores in the area of mathematical problem-solving. While hand-crafted strategies and heuristics are often effective, formulating the problem as a meta-level MDP makes it possible to automate the design of the learning curriculum by discovering optimal planning strategies that can be taught to people through intelligent tutoring systems.

Different approaches to solving fully observable meta-level MDPs exist (e.g. \citep{hay2014selecting,callaway2017learning,consul2021improving,svegliato2018adaptive,Griffiths2019Doing}). While dynamic programming provides an exact solution, the size of environments it can be applied to is severely limited due to the exponentially increasing space of possible belief states. Therefore, approximate methods have been developed. One approach is the meta-greedy policy that always chooses the computation that maximizes the immediate improvement in the plan minus its cost \citep{RUSSELL1991361}, reducing the complexity of planning by only planning ahead for one time step. \citet{callaway2017learning} developed a method that combines multiple approximations of the value of computation (VOC) to estimate the benefit of computations, and \citet{consul2021improving} improved the method's scalability using a hierarchical decomposition. Another recent approach by \citet{ellis2019write} formalized the related problem of program synthesis as an MDP. 


While models of human planning rooted in the fully observable setting assumed that a computation always reveals the true reward of the simulated action \citep{Callaway2022Rational,Griffiths2019Doing}, thinking about an action's outcomes in the real world rarely leads to certainty \citep{vul2014one,guez2012efficient}. Instead, each time a person thinks about an action's outcome, the scenario they imagine might be different, and the more uncertain the outcome is, the more variable the imagined outcomes will be. We incorporate this variability into the meta-level MDP model of human planning \citep{Callaway2022Rational}. In the fully observable meta-level MDP, each computation reveals fully certain information about the object-level environment, which can be used to directly update the belief state. In the partially observable setting, computations instead reveal noisy information about the object-level environment. As a result, the belief state also reflects this uncertainty, and now has to be updated based on uncertain observations instead of reflecting precise beliefs about the object-level environment.

This difference is closely related to classical research on partially observable Markov decision processes (POMDP), an extension to MDPs in which an agent is unable to observe the exact environment state and instead receives uncertain information \citep{kaelbling1998planning}. Formally, the POMDP can be described as a tuple $(\mathcal{S}, \mathcal{A}, P, r, \Omega, O)$, which extends the MDP framework with a set of observations $\Omega$ and an observation function $O(s, a)$ which provides an observation about the environment depending on the current state $s$ and the agent's action $a$ \citep{kaelbling1998planning}. A classic example of POMDPs in the literature is the tiger problem \citep{kaelbling1998planning}, in which an agent has to choose between two doors with a tiger behind one of them. Observations in this environment are noisy, the agent can't see the tiger and has to rely on auditory clues like the tiger's growling. Partial observability in our meta-level MDP model functions analogously to this example, instead of observing the object-level environment directly (e.g. seeing the tiger through a glass door), computations can only reveal noisy information (e.g. listening for sounds).

In our model of planning in partially observable environments, computations produce samples from the agent's probability distribution on what the true reward might be. Each sample is then integrated into the probabilistic belief state using Bayesian belief updates. 
This allows computations to be repeated multiple times, iteratively increasing the accuracy of one's belief. We believe that advancing automatic strategy discovery to partially observable environments is an important stepping stone on the path to leveraging automatic strategy discovery to improve human decision-making in the real world.


To teach planning strategies discovered by solving meta-level MDPs, we build upon existing work known as \textit{cognitive tutors}, in which strategies are taught fully automatically using intelligent tutoring systems \citep{Callaway2021Leveraging}. These tutors teach people an optimal planning strategy in meta-level MDPs by letting users practice in the environment while giving them direct feedback on the planning operations they choose. The feedback on a chosen planning operation consists of the information which computation was optimal and, in case of a mistake, a delay penalty of a few seconds during which the user can't interact with the learning environment. The duration of the delay is proportional to the difference between the meta-level Q-values of the optimal planning operation and the chosen planning operation, causing worse planning actions to trigger longer delays \citep{Callaway2021Leveraging}. This approach differs from past intelligent tutoring systems for teaching metacognitive skills, which usually rely on teaching static, hand-crafted metacognitive strategies \citet{aleven2006toward, chi2010meta, ozsoy2009effect}, sometimes combined with adaptive tutoring systems that model the learner's understanding \citep{guerra2017book, roll2005modeling}. While hand-crafted strategies have been successful in the past, cognitive tutors teaching automatically discovered metacognitive strategies have the advantage of (1) not requiring hand-crafted heuristics as (near-)optimal metacognitive strategies are derived directly from the meta-level MDP model of the planning task, and (2) being able to track the learner's belief state exactly through an experimental paradigm which externalizes the human planning process \citep{callaway2017mouselab}, enabling accurate and situation-dependent feedback.

\section{Discovering effective cognitive strategies for planning in partially observable environments}

Partial observability considerably increases the difficulty of solving the meta-level MDP because the number of possible sequences of computations is no longer limited by the number of nodes. None of the previous methods could handle this challenge. To overcome this problem, we developed the meta-greedy policy for partially observable environments (MGPO), a meta-greedy planning method that selects computations using a myopic approximation of the VOC. The myopic VOC only considers the immediate benefit of computations: how likely it is that the new information will change the agent's plan. 
In the partially observable setting, it is necessary to estimate the probability that the next noisy observation reveals information that will change the agent's plan after integrating the information into the agent's posterior belief state. 
We calculate the probabilities and the expected improvement of these events by applying the rules of probability theory. Using the myopic VOC calculation, the overall planning strategy is constructed by iteratively choosing the computation with the highest VOC until the cost of planning becomes higher than the expected improvement, at which point planning is terminated, and the plan with the highest expected reward is chosen. A major benefit of MGPO is its computational efficiency, which allows computing the approximated VOC of computations in an online fashion, a critical component to building a feedback-based intelligent cognitive tutor.

The goal of strategy discovery methods is to discover a planning strategy that specifies which computation people should select next given the currently known information about the environment. This problem is referred to as the \textit{meta-level} problem, in which actions are stochastic simulations that update the agent's belief about rewards in the \textit{object-level} problem. The object-level problem is deterministic and can be described by a directed acyclic graph $G=(V,E)$. The graph has a start node $v_0$ and a set of goal nodes $V_G\subseteq V$ that have no child nodes. Each instance of the object-level problem is initialized by assigning its nodes $g\in G$ randomly sampled ground truth rewards, which are fixed during the planning process and drawn from a multivariate Normal distribution: $(t_1,...,t_N)\sim(\mathcal{N}(\mu_{1}^{\text{GT}}, \tau_{1}^{\text{GT}}),...,\mathcal{N}(\mu_{N}^{\text{GT}}, \tau_{N}^{\text{GT}}))$ \citep{consul2021improving}. We define $\text{Paths}(G)$ as the set of paths that start at the start node and end at a goal node. The object-level actions then consist of selecting a path to traverse, accumulating the ground-truth rewards of the visited nodes.

\subsection{Formalizing the strategy discovery problem as a meta-level MDP}
Following previous work on strategy discovery in other domains \citep{Griffiths2019Doing,Callaway2022Rational,consul2021improving}, we model the problems of discovering planning strategies for partially observable environments as a meta-level Markov Decision Process $M=(\mathcal{B}, \mathcal{C}, T_{\text{meta}}, r_{\text{meta}})$ \citep{hay2014selecting}, consisting of a belief state $b\in\mathcal{B}$, a set of computations $\mathcal{C}$, a reward function $r_{\text{meta}}$, and transition probabilities $T_{\text{meta}}$. We now define each of the four components of our meta-level MDP formulation of the strategy discovery problem in turn.

\paragraph{Belief states}
In general, the belief state $b\in\mathcal{B}$ of a meta-level MDP represents the agent's current knowledge about the environment's reward. We model the agent's beliefs about the rewards at the $N$ locations of the partially observable environment as a multivariate Normal distribution. The entries in the belief state $b_i=(\mu_i,\tau_i)$ represent the agent's posterior distribution about the reward of node $i\in V$ given the information that has been observed so far. The initial belief state is defined as $b^{(0)}=(\mathcal{N}(\mu_{1}^{\text{GT}}, \tau_{1}^{\text{GT}}),...,\mathcal{N}(\mu_{N}^{\text{GT}}, \tau_{N}^{\text{GT}}))$. This belief reflects the statistics of the environment, in the sense that the ground-truth rewards of each problem instance are drawn from the same distribution.

\paragraph{Meta-level actions (computations)}
The meta-level actions $\{c_1,...,c_n,\perp\}\in\mathcal{C}$ are computations. They comprise planning operations $c_i$ that reveal information about the environment, as well as the termination operation $\perp$ that represents terminating planning and executing the plan with the highest expected reward. Unlike in fully observable environments, in partially observable environments, each meta-level action can be repeated arbitrarily many times.

\paragraph{Meta-level reward function}
All planning operations have a fixed cost $r_{\text{meta}}(b,c_i)=-\lambda$. Termination leads to a reward that is received from moving along the path $p\in \text{Paths}(G)$ with the highest expected reward according to the current belief state (Equation~\ref{eq:best_path} and~\ref{eq:expected_reward}).
\begin{equation}
    p_{\text{max}}(b)=\argmax_{p\in \text{Paths}(g)}\sum_{i\in p}\mu_i
    \label{eq:best_path}
\end{equation}
\begin{equation}
    r_{\text{meta}}(b,\perp)=\sum_{i\in p_{\text{max}}}\text{t}_i
    \label{eq:expected_reward}
\end{equation}

\paragraph{Meta-level transition probabilities}
The transition probabilities $T_{\text{meta}}(b,c,b')$ specify how the belief state is updated by different meta-level actions. Each computation $c_i$ reveals noisy information $o_i\sim\mathcal{N}(t_i,\tau_{\text{obs}})$ about the expected reward of the corresponding node represented by belief state $b_i$. The precision parameter $\tau_{\text{obs}}$ is a global parameter of the environment that specifies how accurate planning operations are. An observation $o_i$ is added to the current belief about that state $b_i$ by computing the posterior mean and precision $(\mu_i,\tau_i)$ of the updated belief state (Equation~\ref{eq:posterior}) using the Gaussian conjugate prior \citep{murphy2007conjugate}.

\begin{equation}
    \left(\mu_i^{t+1},\tau_i^{t+1}\right)=\left(\frac{\tau_i^t\mu_i^t+\tau_{\text{obs}}o_i}{\tau_i^t+\tau_{\text{obs}}},\tau_i^t+\tau_{\text{obs}}\right)
    \label{eq:posterior}
\end{equation}

Together, the initial belief state, the set of computations, the transition probabilities, and the meta-level reward function formalize the problem of discovering resource-rational planning strategies for partially observable environments. We now turn to the \textit{meta-greedy policy for partially observable environments} (MGPO), our algorithm for approximating the optimal policy $\pi^*$. The optimal policy $\pi^*$ is defined as the policy that achieves the highest cumulative meta-level reward over all sequential planning steps $t$ until the termination action $\perp$ is chosen (see Equation~\ref{eq:optimal_policy}).

\begin{equation}
    \pi^* = \arg\max_\pi \mathbb{E}\left[\sum_{t} r_{\text{meta}}\left(b^{(t)}, \pi\left(b^{(t)}\right)\right)\right]
    \label{eq:optimal_policy}
\end{equation}

\subsection{The MGPO algorithm}
To discover resource-rational planning strategies in partially observable environments, we approximate the meta-level Q-value $Q(c_i, b)$ of a planning operation $c_i$ in belief state $b$ using a greedy myopic approximation to the value of computation (VOC) \citep{RUSSELL1991361}. Since in the myopic setting, only actions that change the expected best path $p_{\text{max}}$ have a positive VOC, we evaluate the VOC of operation $c_i$ by calculating the probability that an observation $o_i$ will change $p_{\text{max}}$. Specifically, the VOC is approximated by calculating the probability of changing the expected path times the benefit of changing the expected best path minus the cost of planning.

We define $r_{\text{max}}(b)$ as the expected reward of $p_{\text{max}}$ (Equation~\ref{eq:r_max}), $r_i(b)$ as the expected reward of the expected best path leading through node $i$ (Equation~\ref{eq:r_i}), and $r_{\text{alt}}(b)$ as the expected reward of the expected best path that does not lead through node $i$ (Equation~\ref{eq:r_alt}).

\begin{equation}
    r_{\text{max}}(b)=\max_{p\in \text{Paths}(g)}\sum_{j\in p}\mu_j
    \label{eq:r_max}
\end{equation}

\begin{equation}
    r_i(b)=\max_{(p\in  \text{Paths}(G): i\in p)}\sum_{j\in p}\mu_j
    \label{eq:r_i}
\end{equation}

\begin{equation}
    r_{\text{alt}}(b)=\max_{(p\in  \text{Paths}(G): i\notin p)}\sum_{j\in p}\mu_j
    \label{eq:r_alt}
\end{equation}

There are two scenarios in which an observation $o_i$ of the node $i$ can change the optimal path. If the node $i$ does not lie on the path with the highest expected value ($r_{\text{alt}}=r_{\text{max}}$), the best path can only change through observation $o_i$ if the observed value raises the expected value of $r_i$ above $r_{\text{alt}}$. If the node $i$ lies on the path that currently has the highest expected value ($r_i=r_{\text{max}}$), the best path changes if the observed value lowers the expected value of $r_{\text{max}}$ below the value of the best alternative path $r_{\text{alt}}$. In each case, we can define a threshold value $t$ that specifies an upper/lower bound on how small/large the observation would have to be to change $p_{\text{max}}$. 

For the case where node $i$ does not lie on the optimal path, Equation~\ref{eq:threshold_1} shows how the current observation changes the value of the path through node $i$ from the previous expected value for node $i$ (i.e., $\mu_i$) to the expected value after the observation (i.e., $\mu_i'$; see Equation~\ref{eq:posterior}). If the left-hand side of Equation~\ref{eq:threshold_1} is larger than its right-hand side, the path through $v_i$ now has a higher expected value than the previous best path. Solving for the observation $o_i$ that makes both sides equal gives us the threshold $t$ defined in Equation~\ref{eq:threshold_2}. 

\begin{equation}
    r_i - \mu_i + \frac{\tau_i\mu_i+\tau_{\text{obs}}o_i}{\tau_i+\tau_{\text{obs}}}>r_{\text{alt}}
    \label{eq:threshold_1}
\end{equation}

\begin{equation}
    o_i > t = \left(1+\frac{\tau_i}{\tau_{\text{obs}}}\right)(r_{\text{alt}}-r_i)+\mu_i
    \label{eq:threshold_2}
\end{equation}

We will now present the computations for the first case (node $i$ does not lie on the expected best path) in detail (see line~\ref{alg:v_i_start} to~\ref{alg:v_i_end} of Algorithm~\ref{alg:voc}). For the alternative case (node $i$ lies on the expected best path), the VOC can be computed analogously (see line~\ref{alg:v_alt_start} to~\ref{alg:v_alt_end} of Algorithm~\ref{alg:voc}).

So far, we have calculated the threshold $t$ an observation $o_i$ needs to exceed to change the expected best path. We now move to estimating the potential improvement in the new expected best path through the updated belief state $b_i$. The probability of receiving an observation $o_i\sim\mathcal{N}(\mu_i, \sigma_\text{obs})$ that is higher than $t$ is given by the complement of the cumulative distribution function (CDF) of the Normal distribution (Equation~\ref{eq:p_change}). Assuming the threshold is reached, the expected value of an observation that lies above the threshold can be calculated using the expected value of a lower-bounded truncated Normal distribution \citep{burkardt2014truncated} with its lower bound set to $t$ (see Equation~\ref{eq:o_change}). Using $o_{\text{change}}$ as the expected value of such an observation, we use Equation~\ref{eq:posterior} to compute the conditional posterior expectation $\mu_i'$ given that the observation turns the inspected path into the optimal path. 

To compute the myopic VOC, we calculate the overall expected improvement in $r_{\text{max}}$ (the probability of changing the expected best path $p_{\text{change}}$ times the benefit of changing the expected best path $o_{\text{change}}$) while taking the cost of planning $\lambda$ into account. Equation~\ref{eq:voc} shows the calculation, where the expected improvement ($r_i+\mu_i'-r_{\text{alt}}-\mu_i$) is the difference between the value of the updated path $r_i$ through node $i$ with its updated expected value $\mu_i'$ and the value of the previous best path $r_{\text{alt}}$. The expected improvement is multiplied by the probability of observing an improvement (i.e., $p_{\text{change}}$; see Equation~\ref{eq:p_change}) and the cost of the planning operation is subtracted to arrive at the myopic VOC. 

Since our VOC calculation is purely myopic, it can underestimate the VOC due to not taking the value of additional future information into account. This is especially problematic when deciding whether to terminate planning (i.e. whether the VOC of the expected best computation exceeds the cost of planning $\lambda$). To circumvent the issue of terminating planning too early, we introduce a new hyperparameter $w_\lambda\in[0,1]$, which can be used to adjust the amount of planning before the algorithm terminates by scaling the (imagined) cost of planning in the VOC calculation. We optimize $w_\lambda$ using 50 steps of Bayesian optimization \citep{mockus2012bayesian}.

\begin{equation}
    p_{\text{change}}=1-\Phi\left(\frac{t-\mu_i}{\sigma_{\text{obs}}}\right)
    \label{eq:p_change}
\end{equation}

\begin{equation}
    o_{\text{change}}=\mu_i+\sigma_{\text{obs}}\frac{\phi\left(\frac{t-\mu_i}{\sigma_{\text{obs}}}\right)}{1-\Phi\left(\frac{t-\mu_i}{\sigma_{\text{obs}}}\right)}
    \label{eq:o_change}
\end{equation}

\begin{equation}
    \hat{voc}(c_i, b) = (1-w_\lambda)(p_{\text{change}})(r_i - r_{\text{alt}} - \mu_i + \mu_i') - w_\lambda\lambda
    \label{eq:voc}
\end{equation}


Lines~\ref{alg:mgpo_start} to~\ref{alg:mgpo_end} of Algorithm~\ref{alg:voc} describe the complete MGPO algorithm, which consists of computing the VOC for all available planning actions $c_i$ and selecting the planning action with the highest VOC $c_\text{max}$. This process is repeated iteratively until no planning action with a positive VOC is found, at which point the termination action $\perp$ is selected to stop the planning process.

\begin{algorithm}
\caption{MGPO algorithm including the myopic VOC calculation}\label{alg:voc}
\begin{algorithmic}[1]
\Function{myopic\_voc}{$c_i$, $b$} 
\State $r_i \gets \max_{(p\in  \text{Paths}(G)|v_i\in p)}\sum_{v_i\in p}\mu_i$ \Comment{Best path including node $v_i$}
\State $r_{\text{alt}} \gets \max_{(p\in  \text{Paths}(G)|v_i\notin p)}\sum_{v_i\in p}\mu_i$  \Comment{Best alternative path}
\State $t \gets (1+\frac{\tau_i}{\tau_{\text{obs}}})(r_{\text{alt}}-r_i)+\mu_i$ \Comment{Observation threshold}
\State $\sigma_{\text{obs}} \gets \sqrt{(\frac{1}{\sqrt{\tau_{\text{obs}}}})^2+(\frac{1}{\sqrt{\tau_{i}}})^2}$ \Comment{Combined uncertainty in observation} \label{alg:sigma_comp}
\If{$r_i>r_{\text{alt}}$} \label{alg:v_i_start} \Comment{Investigated node is on best path}
    \State $p_{\text{change}} \gets \Phi(\frac{t-\mu_i}{\sigma_{\text{obs}}})$ \Comment{Probability of changing the optimal path}
    \State $o_{\text{change}} \gets \mu_i - \sigma_{\text{obs}}\frac{\phi(\frac{t-\mu_i}{\sigma_{\text{obs}}})}{\Phi(\frac{t-\mu_i}{\sigma_{\text{obs}}})}$ \Comment{Conditional expected observation}
    \State $\mu_i' \gets \frac{\mu_i\tau_i+{o_{\text{change}}}\tau_{\text{obs}}}{\tau_i+\tau_{\text{obs}}}$ \Comment{Updated belief state $\mu_i$}
    \State voc $\gets (p_{\text{change}})(r_{\text{alt}} - r_i + \mu_i - \mu_i')$ \label{alg:v_i_end} \Comment{Expected value of computation}
\ElsIf{$r_i<=r_{\text{alt}}$} \label{alg:v_alt_start} \Comment{Investigated node is not on best path}
    \State $p_{\text{change}} \gets 1-\Phi(\frac{t-\mu_i}{\sigma_{\text{obs}}})$ \Comment{Probability of changing the optimal path}
    \State $o_{\text{change}} \gets \mu_i + \sigma_{\text{obs}}\frac{\phi(\frac{t-\mu_i}{\sigma_{\text{obs}}})}{1-\Phi(\frac{t-\mu_i}{\sigma_{\text{obs}}})}$ \Comment{Conditional expected observation}
    \State $\mu_i' \gets \frac{\mu_i\tau_i+{o_{\text{change}}}\tau_{\text{obs}}}{\tau_i+\tau_{\text{obs}}}$\Comment{Updated belief state $\mu_i$}
    \State voc $\gets (p_{\text{change}})(r_i - r_{\text{alt}} - \mu_i + \mu_i')$ \label{alg:voc_comp} \label{alg:v_alt_end} \Comment{Expected value of computation}
\EndIf
\State \Return $(1-w_\lambda)$voc - $w_\lambda\lambda$ \Comment{Final VOC with cost adjustment}
\EndFunction
\vspace{1em}
\Function{mgpo}{$b$} \label{alg:mgpo_start}
\State $c_{\text{max}} \gets \max_i\textsc{MYOPIC\_VOC}(c_i, b)$ \Comment{Select computation with highest VOC}
\If{$c_{\text{max}}>0$}
    \State \Return $c_{\text{max}}$
\ElsIf{$c_{\text{max}}<=0$}
    \State \Return $\perp$ \Comment{Stop planning if no computation is beneficial}
\EndIf
\EndFunction \label{alg:mgpo_end}
\end{algorithmic}
\end{algorithm}

\section{Evaluating our strategy discovery method in simulations} 
To evaluate how resource-rational the strategy discovered by MGPO is, we compared its resource-rationality against two baseline methods in two simulation experiments.

\subsection{Benchmark problems: strategy discovery for partially observable environments}
\begin{figure}[ht]
  \centering
  \includegraphics[width=\textwidth, trim = 0cm 17.5cm 0cm 0cm, clip]{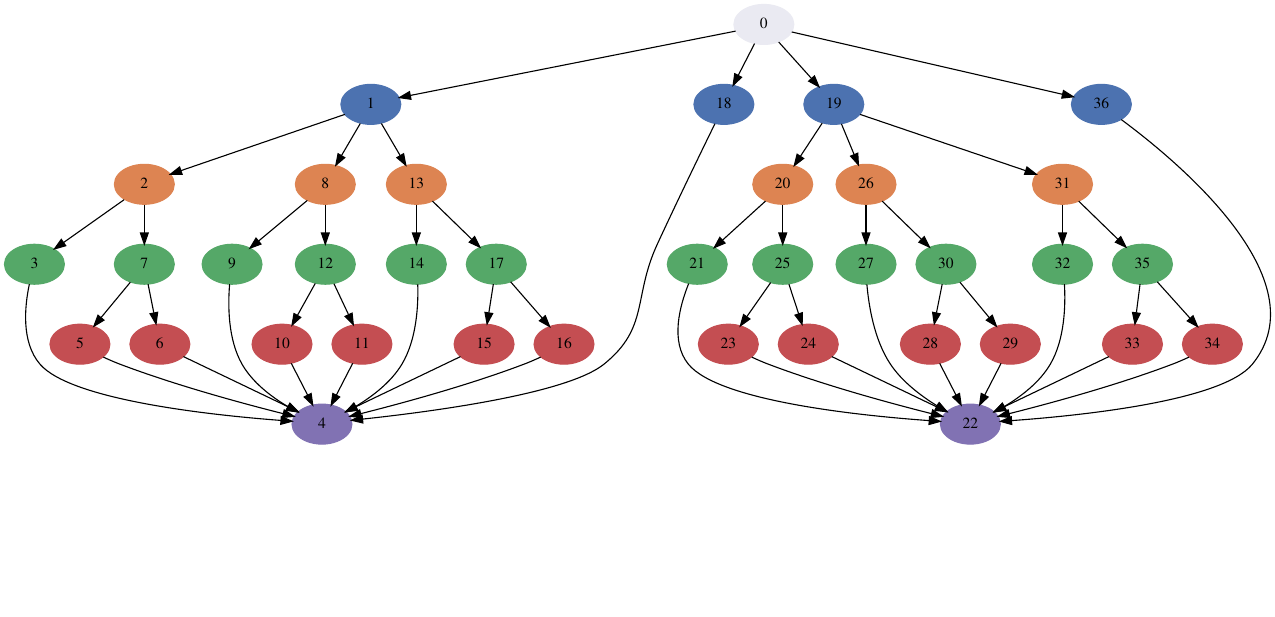}
  \caption{Environment used in the simulation evaluation. The reward of each node is independently sampled from a Normal distribution: $\mathcal{N}(0,5)$ for nodes colored in blue, $\mathcal{N}(0,10)$ for orange nodes, $\mathcal{N}(0,20)$ for green nodes, $\mathcal{N}(0,40)$ for red nodes, $\mathcal{N}(0,100)$ for node 4, and $\mathcal{N}(0,120)$ for node 22.}
  \label{fig:env_sim}
\end{figure}

The simulation experiments are run on 4 benchmark environments of different sizes introduced by \citet{consul2021improving}. The environments were built from 2 to 5 blocks of 18 nodes, with each block leading to a different goal node. Including the root node, environments therefore had a total size of at least 37 and up to 91 nodes. The environments all follow an increasing variance structure, where the uncertainty about the reward of the node increases the further away it is from the starting node (see Figure~\ref{fig:env_sim}). The rewards for each environment instance were initialized randomly following an increasing-variance reward structure, where nodes closer to the leaves had a higher variance in their reward (see Figure~\ref{fig:env_sim} for an example of the smallest evaluation environment and an explanation of the reward structure). To make the environment partially observable, the observations generated by the planning operations were randomly sampled from Normal distributions centered around the environment's ground truth values with a fixed precision parameter $\tau=0.005$. The cost of computations was fixed to $\lambda=1$ (matching the previous experiment by \citet{consul2021improving}) for the first simulation experiment, and $\lambda=0.05$ for the second simulation experiment. All methods were evaluated by their average meta-level return (i.e. their expected termination reward) on the same 5000 randomly generated environment instances. 

\subsection{Baseline methods} 

For our baseline methods, we selected a discretized version of the myopic VOC as used by \citet{consul2021improving} and PO-UCT \citep{silver2010monte}, a sample-based planning algorithm utilizing Monte-Carlo tree search that is suitable for partially observable environments. We selected the discretized myopic VOC method because it is the closest established method for solving meta-level MDPs that we could apply to the partially observable setting without major modifications. While PO-UCT has not been applied to meta-level MDPs yet, we chose the method to compare MGPO to an established search-based method for solving POMDPs. We opted for PO-UCT over POMCP, the full method described by \citet{silver2010monte}, as our environment allows us to compute exact belief state updates.

\paragraph{Discretization}
Our first baseline method approximates the meta-greedy policy \citep{RUSSELL1991361} using discretization. Following \citet{consul2021improving}, when calculating the VOC, we discretized the continuous observation that each planning operation generated into a categorical variable with four possible values based on the current belief of the expected mean $\mu$ and variance $\sigma$ for each node $i$: ($\mu_i-2\sigma_i, \mu_i-\frac{2}{3}\sigma_i, \mu_i+\frac{2}{3}\sigma_i, \mu_i+2\sigma_i$). We calculated the probability that the observation would fall into a given bin using the Normal distribution's cumulative distribution function. Using these 4 possible outcomes for every planning operation, the action with the highest immediate increase in expected reward is chosen while assuming that planning will terminate after the selected planning operation has been carried out: 
\begin{equation}
    c=\arg\max_{c\in \mathcal{C}}\sum_{(p, b')\in \text{disc}(T(b,c,\cdot))}(p)(r(b',\perp)-r(b,\perp)+r(b,c))
    \label{eq:myopic_voc}
\end{equation}
where $\text{disc}(T(b,c,\cdot))$ is the vector containing the probabilities and posterior belief states of the transitions that can result from witnessing the four possible values of discretized observation.

\begin{table}[ht]
\small
\centering
\makebox[\textwidth]{
\begin{tabular}{ll|rrrr|rrrr}
\toprule
     \multicolumn{2}{r}{Environment (Goals)} & \multicolumn{1}{c}{2} &\multicolumn{1}{c}{3} & \multicolumn{1}{c}{4} &   \multicolumn{1}{c}{5} & \multicolumn{1}{c}{2} &\multicolumn{1}{c}{3} & \multicolumn{1}{c}{4} &   \multicolumn{1}{c}{5} \\
    \midrule
Cost & \multicolumn{1}{l}{Steps} &\multicolumn{4}{c}{Exploration Coefficient} & \multicolumn{4}{c}{Rollout Depth}   \\
\midrule
0.05 & 10   &            100.0 &  100.0 &  100.0 &  100.0 &          0.0 &  3.0 &  3.0 &  0.0 \\
     & 100  &            100.0 &  100.0 &   10.0 &    5.0 &          3.0 &  3.0 &  3.0 &  3.0 \\
     & 1000 &              1.0 &    5.0 &   10.0 &  100.0 &          3.0 &  3.0 &  3.0 &  3.0 \\
     & 5000 &              5.0 &   50.0 &    5.0 &    5.0 &          0.0 &  0.0 &  3.0 &  3.0 \\
1.00 & 10   &            100.0 &  100.0 &  100.0 &  100.0 &          0.0 &  3.0 &  3.0 &  0.0 \\
     & 100  &             10.0 &  100.0 &    5.0 &   50.0 &          0.0 &  0.0 &  0.0 &  0.0 \\
     & 1000 &            100.0 &   10.0 &   50.0 &  100.0 &          0.0 &  0.0 &  0.0 &  0.0 \\
     & 5000 &              5.0 &  100.0 &  100.0 &   50.0 &          0.0 &  0.0 &  0.0 &  0.0 \\
\bottomrule
\end{tabular}
}
\caption{Exploration coefficient and rollout depth for PO-UCT simulations selected by a hyperparameter grid search.}
\label{table:pouct_sim}
\end{table}

\paragraph{Monte-Carlo Tree Search}
Secondly, we also evaluated our method against the Monte-Carlo tree search algorithm PO-UCT \citep{silver2010monte}. Our implementation used the same discretized state representation with 4 bins per node when running the Monte-Carlo simulations as described in the discretized approximation of the meta-greedy policy. We discretized the simulated belief updates during the sample-based simulations to allow the tree search planning to reach states multiple times and plan multiple steps into the future. 
We evaluated 4 different versions of the algorithm that have different numbers of evaluation steps per action (10 steps, 100 steps, 1000 steps, 5000 steps), of which the lowest was set to roughly match the computation time of our method. While traversing the search tree, nodes are selected using UCB1 \citep{silver2010monte}. When a leaf node is reached, a rollout policy is used to estimate the node's value. Our rollout policy consists of performing a fixed number of uniformly random computations $c\in \mathcal{C}$, after which the termination action $\perp$ is selected. We ran a hyperparameter search over 500 training environments with the rollout depth set to either $0$ or $3$ and exploration coefficients set to one of $\{0.5, 1, 5, 10, 50, 100\}$. For each computational budget used in the simulation experiment, we selected the parameter settings that achieved the highest resource-rationality score and evaluated PO-UCT with the selected parameters on the 5000 test environments. The selected parameters for each environment and cost configuration are listed in Table~\ref{table:pouct_sim}.

\subsection{Results} \label{sec:simulation}

\begin{table}[ht]
\small
\centering
\makebox[\textwidth]{
\begin{tabular}{llrrrr}
\toprule
     & Environment (Goals) &    2 &    3 &    4 &    5 \\
Cost & Algorithm &         &         &         &         \\
\midrule
0.05 & MGPO (ours) &  \textbf{118.97} &  \textbf{158.28} &  \textbf{191.14} & \textbf{223.84} \\ 
     & Meta-greedy policy &  115.47 &  154.33 &  186.91 &  218.45 \\
     & PO-UCT 10 steps &   55.00 &   55.00 &   55.03 &   55.20 \\
     & PO-UCT 100 steps &  114.94 &  152.42 &  186.17 &  218.75 \\
     & PO-UCT 1000 steps &  115.46 &  155.96 &  188.76 &  220.57 \\
     & PO-UCT 5000 steps &  117.41 &  156.53 &  189.51 &  220.94 \\

1.00 & MGPO (ours) &  \textbf{104.75} &  \textbf{142.27} &  \textbf{173.75} &  \textbf{205.27} \\
     & Meta-greedy policy &  103.69 &  141.46 &  173.23 &  204.27 \\
     & PO-UCT 10 steps & -135.95 & -135.67 & -135.89 & -135.64 \\
     & PO-UCT 100 steps &   98.11 &  133.79 &  160.86 &  189.61 \\
     & PO-UCT 1000 steps &  102.23 &  138.33 &  171.64 &  202.69 \\
     & PO-UCT 5000 steps &  100.66 &  140.10 &  172.20 &  203.21 \\
\bottomrule
\end{tabular}
}
\caption{Mean resource rationality score for both cost settings and all four environments. Results are averaged over 5000 environment instances. The PO-UCT versions differ by the number of simulation steps given for each computation.}
\label{table:res_sim}
\end{table}

Table~\ref{table:res_sim} shows each method's average resource rationality scores (\textit{RR-score}). We analyzed the results using a three-way ANOVA to determine the main effect of the algorithm on the \textit{RR-score} for different planning costs and different sizes of the environment. The ANOVA revealed that some methods tended to discover significantly more resource rational planning strategies than others ($F(5, 239952)=32682.5$, $p<.001$). 
A Tukey-HSD post-hoc comparison showed that MGPO performs significantly better than the meta-greedy policy ($p=.001$), the PO-UCT algorithm with 5000 steps ($p=.008$), and all PO-UCT algorithm configurations with less than 5000 steps (all $p<.001$).

On average, MGPO took $0.03$ seconds to select the next computation. The only baseline method with a comparable runtime is PO-UCT with 10 steps ($0.02$~seconds). However, with only 10 computation steps, PO-UCT fails to perform any meaningful planning and performs much worse than all other tested methods (see Table~\ref{table:res_sim}). The other baseline methods required a computational budget more than a 100 times higher than the budget used by the MGPO policy: the meta-greedy policy required $4.89$ seconds, PO-UCT with 100 steps $0.22$ seconds, PO-UCT with 1000 steps $3.22$ seconds, and PO-UCT with 5000 steps $14.19$ seconds.

In summary, our results show that our method is consistently better than alternative methods with a similar computational budget and much faster. Additional analysis of the interaction effects can be found in the supplementary information.

\section{An intelligent tutor for planning in partially observable environments}

\begin{figure}[ht]
     \centering
     \begin{subfigure}[t]{0.49\textwidth}
         \centering
        \includegraphics[width=\textwidth]{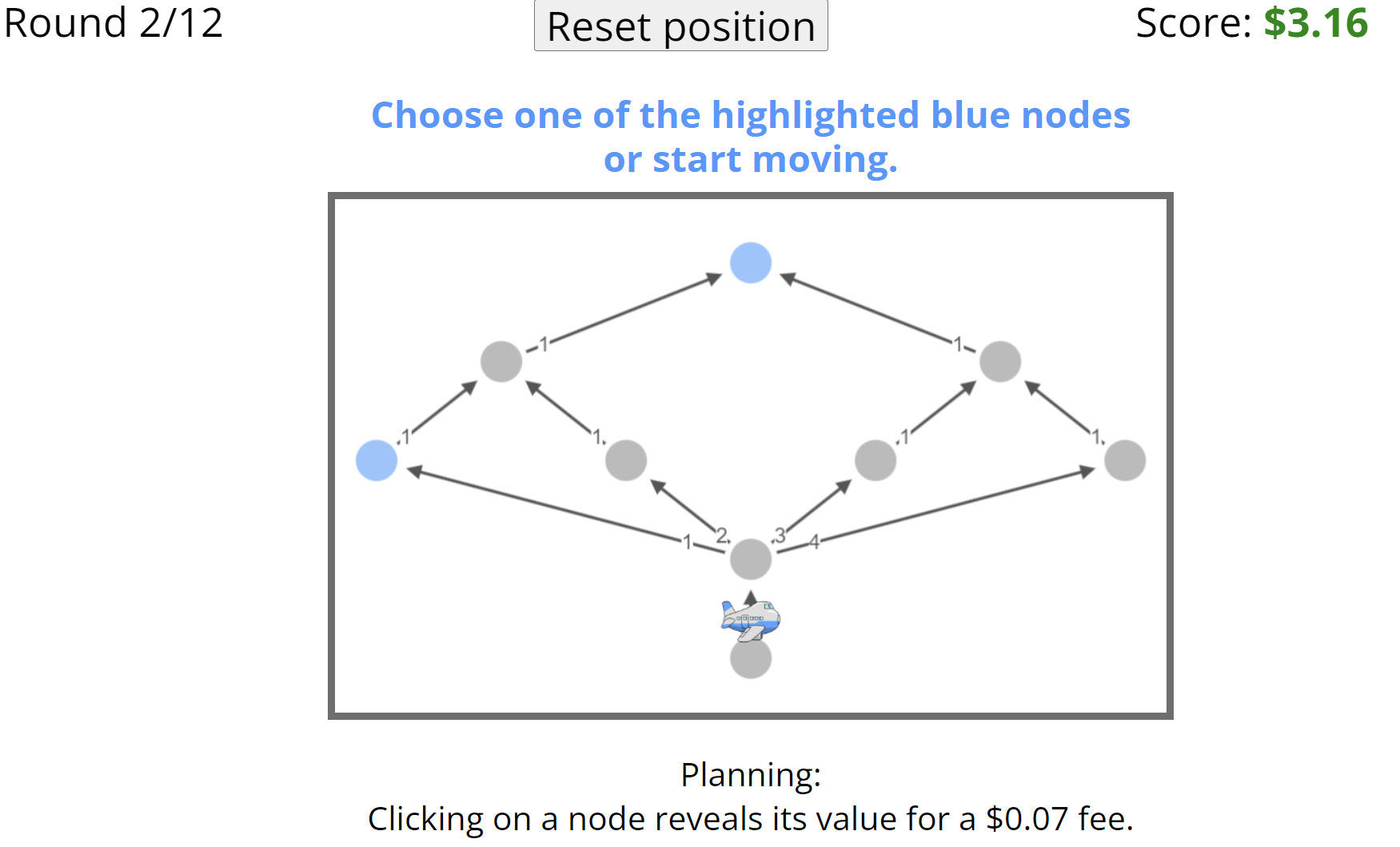}
         \caption{The tutor offers participants a choice between two computations.}
         \label{fig:choice_tutor_left}
     \end{subfigure}
     \hfill
     \begin{subfigure}[t]{0.49\textwidth}
         \centering
         \includegraphics[width=\textwidth]{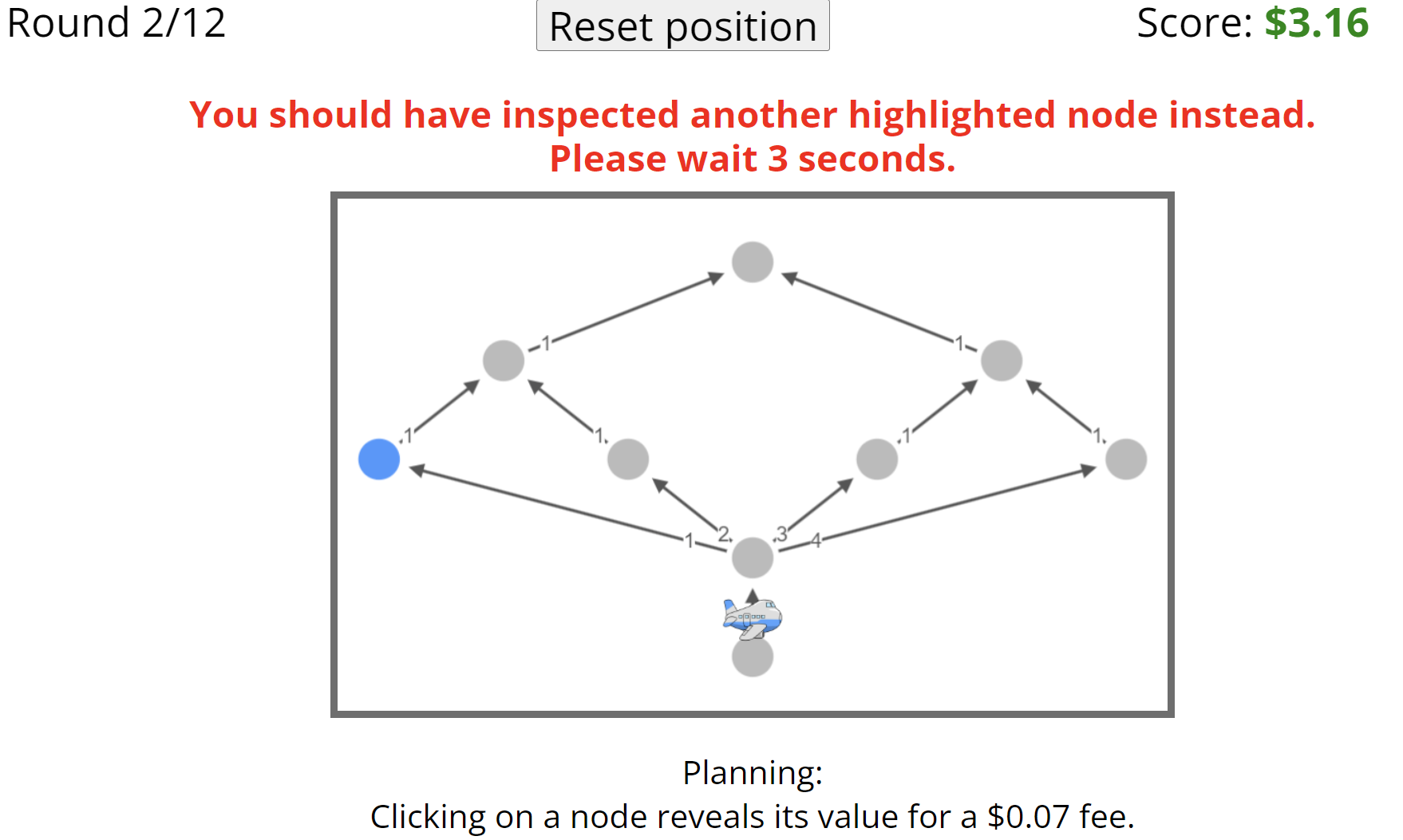}
         \caption{After selecting an incorrect computation, the tutor gives feedback and highlights the correct choice.}
         \label{fig:choice_tutor_right}
     \end{subfigure}
    \caption{Example of the intelligent tutor in the simplified 8-node environment.}
    \label{fig:choice_tutor}
\end{figure}

To teach people the planning strategies discovered by MGPO, we developed a cognitive tutor based on \citet{Callaway2021Leveraging}. Figure~\ref{fig:choice_tutor} shows our tutor, which teaches MGPO's strategy in interactive online sessions. In our experimental environment, \textit{flight planning}, learners are tasked to decide an optimal route for a fictional airplane \citep{consul2021improving}. To decide on a path, learners can investigate individual nodes (airports) by clicking on them to reveal uncertain information about their profitability, which is modeled as a partially observable meta-level MDP. When practicing with the tutor, learners can choose between highlighted planning actions (Figure~\ref{fig:choice_tutor_left}). Upon making a mistake, the tutor provides feedback in addition to a small delay penalty (Figure~\ref{fig:choice_tutor_right}). The design of our tutor is detailed in Section~\ref{sec:ChoiceTutor}. Section~\ref{sec:human_exp} contains a human training experiment in which we demonstrate our tutor's ability to teach planning strategies in the \textit{flight planning} environment.

\subsection{Developing an intelligent tutor that teaches planning in large, partially observable environments} \label{sec:ChoiceTutor}

In this section, we explain the functionality of our intelligent cognitive tutor, starting with the computation of metacognitive feedback in partially observable environments, and then detailing our learning curriculum inspired by Skinner's shaping method \citep{skinner1953shaping}.

\subsubsection{Giving metacognitive feedback on how people plan in partially observable environments} \label{sec:feedback}
To teach planning strategies to people, we adapted the cognitive tutor introduced by  \citep{Callaway2021Leveraging}. Given the optimal strategy for a meta-level MDP, the tutor developed by \citep{Callaway2021Leveraging} teaches the strategy to humans in an interactive learning session by providing them meta-cognitive feedback on their chosen planning actions. If they select a suboptimal planning action, the tutor offers feedback on the correct choice and issues a delay penalty, a short period of time in which the participant has to wait and can't interact with the learning environment. The duration of the delay penalty depends on the difference in VOC between the chosen planning action and the optimal planning, with larger mistakes resulting in longer delays. While the tutor by  \citet{Callaway2021Leveraging} proved effective in small metareasoning problems, its applicability to larger environments is limited since their tutor used dynamic programming to compute the exact Q-function of the meta-level MDP and relied on cached Q-values for all possible belief states. Moreover, their tutor cannot handle partially observable environments. 


We applied our MGPO algorithm to improve the cognitive tutor introduced by \citep{Callaway2021Leveraging} so that it can discover and teach good planning strategies for environments that are large and partially observable. 
Since the VOC values computed by our MGPO algorithm are only approximate, we decided to impose the same delay penalty ($d(c_i)$) for all planning operations $c_i$ who have a negative VOC or whose VOC differs by more than a fixed threshold of $0.05$ from the action with the highest VOC according to MGPO. Choosing the termination action $\perp$ when MGPO does not terminate planning leads to an additional delay penalty that was proportional to an estimate of how valuable it would have been to do more planning in the current belief state. 
The termination delay penalty is calculated by taking the proportion between the VOC of the best meta-action in the current belief state and the highest VOC observed in any earlier planning step (see Equation~\ref{eq:delay}). Intuitively, this proportion describes how valuable planning in the current belief state is when compared to the initial belief state, where the delay for falsely terminating planning is set to $d_{\text{max}}$. In our implementation, we set $d_c$ to $3$ seconds and $d_{\text{max}}$ to $4$ seconds. 

Since planning actions in the partially observable setting are repeatable, the space of possible belief states is extremely vast and pre-computing the tutor's feedback is not feasible. Instead, our tutor provides feedback by using MGPO to compute the VOC of all available planning actions on the fly for every belief state the learner finds themselves in. This is made possible by our simplifications to the delay calculation and the overall computational efficiency of MGPO.

\begin{equation}
    d\left(\perp, b^{(i)}\right) = d_c + d_{\text{max}}\frac{\max_{c\in C}\myopicvoc\left(c, b^{(i)}\right)}{\max_{c\in C, j\in \left[0,i\right)}\myopicvoc\left(c, b^{(j)}\right)}
    \label{eq:delay}
\end{equation}


\subsubsection{A shaping method for teaching cognitive strategies} \label{sec:shaping}


To further help people learn planning strategies, we introduce an improved teaching methodology and training schedule to the intelligent tutor. The cognitive tutors developed by \citep{CognitiveTutorsRLDM, Callaway2021Leveraging} let the learner freely choose their next planning operation as they practice. This causes a potential problem in large environments: the more actions the learner can choose from, the more attempts it will take them to stumble upon the optimal planning operation, and the more errors and delay penalties they must endure along the way. For environments with many possible planning operations, the learning process can thus become very slow and very frustrating. 

To make the intelligent tutor more suitable for teaching planning in large environments, we leverage the method of successive approximations developed in research on animal learning \citep{skinner1953shaping}. That is, the training starts with a minimal version of the planning task, and as the training progresses, the planning task becomes increasingly more similar to the scenario in which people's planning strategies shall be improved. This gradual approximation of a complex planning problem proceeds along two dimensions: the number of planning operations the learner can choose between and the size of the environment. 

Concretely, we created three smaller versions of the full 60-node environment (see Figure~\ref{fig:env_exp}) and increased the number of planning operations the learner is asked to choose between in tandem with the environment size: an 8-node environment with 1 goal node and 2 choices, a 16-node environment with 2 goal nodes and 3 choices, and a 30 node environment with 2 goal nodes and 4 choices  (see Figure~\ref{fig:choice_tutor} for an example of the 8-node environment). 
To offer meaningful choices (i.e., choices with different VOC values), the choices presented to the learner are selected by grouping available computations by their VOC value. The set of choices always includes the planning operation with the highest VOC. The remaining choices are filled up by iteratively adding a computation with a VOC value different from the VOC values of the computations already included in the choice set, uniformly at random. 
In addition to selecting one of the offered choices, participants were also always given the option to terminate planning. To ensure that participants experienced the complete planning strategy, choosing this option prematurely did not actually lead to termination when MGPO's strategy did not terminate planning itself and instead only resulted in a delay penalty as described in Section~\ref{sec:feedback}.
Besides training trials with feedback on the selected choices, the intelligent tutor also demonstrated the taught strategy through step-by-step demonstrations, as used in prior work \cite{consul2021improving}. The demonstrations were created by repeatedly performing the computation selected by MGPO's planning strategy. 

\subsection{Assessing the intelligent tutor in an experiment}\label{sec:human_exp}

Based on the superior performance and computational speed of MGPO, we evaluated the intelligent tutor in an online experiment. In the experiment, we tested if we can improve human resource-rationality by teaching them MGPO's strategy in interactive training sessions. To implement the experiment, we used the Mouselab-MDP paradigm \citep{callaway2017mouselab,Callaway2022Rational,Jain2022}, in which the internal human planning process is externalized through information retrieval actions.

\paragraph{Participants}
We recruited 330 participants through Prolific (\url{www.prolific.co}), of which 37 were excluded through predefined exclusion criteria (for details, refer to the study's pre-registration: \url{https://aspredicted.org/RL3_YDD}). The average age was $28.2$ and of the 330 participants, 164 identified as female. All participants were over 18 years old, fluent in the English language, and gave informed consent to participate in the experiment. Participants were paid $3.80\pounds$ and a performance dependent bonus of up to $2\pounds$ (average bonus $0.91\pounds$). The median duration of the experiment was $40.48$ minutes, resulting in an overall average wage of $6.98\pounds$ per hour. 

\paragraph{Procedure}
Participants were randomly assigned one of three conditions: the experimental condition (\textit{Choice Tutor}) where participants trained with an intelligent tutor teaching MGPO's strategy 
\footnote{The version of MGPO used in the experiment was not the most recent version of our method. Its VOC calculation differed from the current version in two ways. First, the sigma value for observations was calculated taking only the precision of observations into account: $\sigma_{\text{obs}} \gets \frac{1}{\sqrt{\tau_{\text{obs}}}}$  (cf. Line~\ref{alg:sigma_comp} of Algorithm~\ref{alg:voc}). Second, the returned VOC value was calculated without optimizing the cost weight parameter: $w_\lambda$ (cf. Line~\ref{alg:voc_comp} of Algorithm~\ref{alg:voc}). These discrepancies make the strategy taught by our tutor inferior to the strategy discovered by the current version of MGPO by an average difference in resource-rationality score of $5.61$ (evaluated over 500 instances of the 60 node environment, a precision of $0.005$, and a cost value of $0.05$). Therefore, if we had taught participants the more resource-rational strategy discovered by the newer version of MGPO, the benefit of the intelligent tutor would potentially have been even larger. This makes the results of the experiment a lower bound on the effectiveness of the intelligent tutor.},
a control condition without a tutor (\textit{No Tutor}), and a control condition with a \textit{Dummy Tutor}. Participants first received instructions explaining the flight planning task; these instructions were identical across all conditions. Then, participants underwent 12 rounds of training. The training trials utilized the smaller environments described in Section~\ref{sec:shaping}, slowly building up to the full problem complexity: the first 3 training trials used the 8-node environment, trials 4 to 6 used the 16-node environment, trials 7-9 used the 30-node environment, and the final 3 training trials used the full 60-node environment. The type of training depended on the assigned condition: participants of the \textit{Choice Tutor} and \textit{Dummy Tutor} conditions received one demonstration and two feedback trial for each environment size, while participants in the \textit{No Tutor} condition practiced unassisted in three practice trials for each environment size. After completing the training section, participants of all conditions were evaluated in 10 unassisted test trials in the full 60-node environment. Each training and test trial was initialized with randomly sampled ground truth rewards.

\begin{figure}[ht]
  \centering
  \includegraphics[width=\textwidth, trim = 0cm 17.5cm 0cm 0cm, clip]{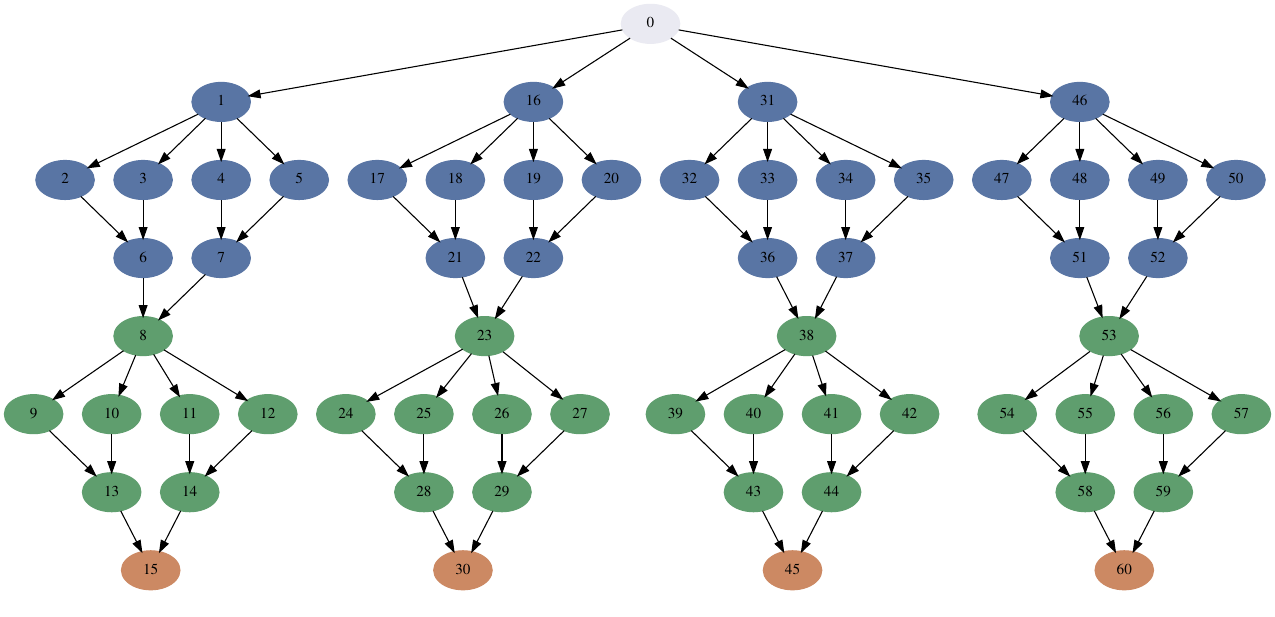}
  \caption{The environment used in the online experiment. Rewards are sampled from Normal distributions depending on the node: $\mathcal{N}(0,5)$ for nodes close to the root (blue), $\mathcal{N}(0,10)$ for intermediate nodes (green), and $\mathcal{N}(0,20)$ for nodes furthest from the root (orange).}
  \label{fig:env_exp}
\end{figure}

\paragraph{Materials}
The evaluation took place in the flight planning scenario introduced by \citet{consul2021improving}. In this task, participants plan the route of an airplane over a network of 60 airports (see Figure~\ref{fig:env_exp}). We adapted the scenario to a partially observable setting. Concretely, in our partially observable version, inspecting an airport does not reveal the true cost or reward of traversing it; instead, it generates a stochastic  observation of what that reward might be. Each observation was drawn from a list of 200 precomputed samples from the observation distribution of the corresponding meta-level MDP. To make the task manageable for people, the posterior means of the airports' rewards were shown to the participant and updated with each observation. To make the performance of the three groups as comparable as possible, all three conditions used the same set of precomputed random observations and environment rewards. Furthermore, we changed the environment's reward distributions to the increasing variance structure described in Figure~\ref{fig:env_exp}.

The cost of clicking and the precision of observations were randomized across participants. The cost values were sampled from a Normal distribution $\lambda\sim\mathcal{N}(0.05, 0.002)$ and clipped to $[0.01,0.5]$, and precision values were sampled from a Normal distribution $\tau\sim\mathcal{N}(0.005, 0.002)$ and clipped to $[0.0001, 0.1]$. These parameters and clipping ranges were chosen manually to ensure a reasonable amount of planning in the resulting optimal strategies. A set of 100 parameter combinations was pre-generated to allow generating observation samples in advance and ensure a comparable distribution of parameters between conditions.

Participants in the \textit{Choice tutor} condition were trained by the intelligent cognitive tutor described in Section~\ref{sec:ChoiceTutor}. 
The Dummy Tutor was designed to reveal similar information about the environment's structure as the Choice Tutor without revealing the strategy taught by the Choice Tutor. To achieve this, the Dummy Tutor differed from the Choice Tutor in three ways. First, whereas the Choice Tutor always includes the optimal planning operation in the choice set, the Dummy Tutor asks the learner to choose from a randomly generated set of planning operations. Concretely, the Dummy Tutor first randomly selects the distance from the starting position (e.g., 3 steps) and then randomly selects two airports that are that far away from the starting position.
Second, when video demonstrations are shown to the participant, the meta-level actions are selected at random. In the video demonstrations and practice trials of the dummy tutor condition, the overall number of meta-level actions was fixed to the number of meta-level actions performed by MGPO's strategy. Third, unlike in the Choice Tutor condition, the number of choices offered by the Dummy Tutor was kept at 2 throughout training.
The reason was that because participants in the Dummy Tutor condition received random feedback, they cannot be expected to master even the simplified selection between only two choices.

\paragraph{Data analysis}
We calculated the five performance measures (i.e., \textit{RR-score}, \textit{click agreement}, \textit{termination agreement}, \textit{repeat agreement}, and \textit{goal planning}) as follows. 
The \textit{RR-score} measures how resource-rational a planning strategy is by computing the expected value of executing a plan while taking the cost of the used planning operations into account. The participant's \textit{RR-score} was calculated by adding the posterior expectation given the participant's observations of all nodes on the chosen object-level path and subtracting the total cost of all performed meta-level actions. 
The \textit{click agreement} score quantifies how similar to MGPO's strategy the participant's chosen meta-level actions are. This measure is computed for each trial by taking the proportion of meta-level actions chosen by the participant that match the meta-level action chosen by MGPO in the same belief state. 
Similarly, \textit{repeat agreement} measures to which extent the participant's strategy and MGPO's strategy agreed on whether to collect more information on an already inspected node versus inspecting a new node. We defined this measure as the proportion of the participant's meta-level actions that either repeated a meta-level action when MGPO's strategy would also perform a (potentially different) repeated meta-level action, or performed a non-repeated meta-level action when MGPO's strategy would also perform a (potentially different) not repeated meta-level action. We considered a meta-level action as repeated if it had been chosen before at any earlier time step. 
Lastly, \textit{termination agreement} measures if participants learned when to terminate selecting meta-level actions if and only if the strategy discovered by MGPO would terminate planning. This measure was calculated using a balanced accuracy measure \citep{brodersen2010balanced} where true positive cases are defined as terminating planning when MGPO's strategy also terminates planning; true negative cases are defined as not terminating when MGPO's strategy also did not terminate planning; false positive cases are defined as terminating planning when MGPO's strategy continued to plan; and false negative cases are defined as not terminating planning when MGPO's strategy terminated planning. 
We also investigated if participants learned to follow a general \textit{goal planning} strategy of first examining nodes furthest away from the root node and then planning backwards, which is a component of MGPO's strategy. For each trial, we calculated whether a participant followed the \textit{goal planning} strategy by testing whether their selected meta-level actions are consistent with the increasing variance reward structure of the environment (see Figure~\ref{fig:env_exp}). Specifically, we tested whether participants performed at least one meta-level action in the nodes with the highest variance (goal nodes) before investigating a node with a lower variance, and whether participants investigated at least one node with a medium variance before investigating a node with a low variance. Participants were considered to have learned this goal-directed planning strategy if they applied it in more than half of the test trials.

We analyzed the main differences between conditions for the \textit{RR-score}, \textit{click agreement}, \textit{repeat agreement}, and \textit{termination agreement} using a robust ANOVA-type statistic with Box approximation \citep{box1954some} implemented in the nparLD R package \citep{nparld} and used additional ANOVA-type statistics for pairwise post hoc comparisons. We used z-tests to compare the proportion of participants who learned to follow the \textit{goal planning} strategy and corrected for multiple comparisons using the Benjamini-Hochberg method \citep{benjamini1995controlling}.

\subsection{Results} \label{exp_results}

\begin{table}[ht]
\small
\centering
\makebox[\textwidth]{
\begin{tabular}{l|rrrrrrrrrr}
\toprule
{} & \multicolumn{1}{c}{\textit{Goal}} & \multicolumn{2}{c}{\textit{RR-}} & \multicolumn{2}{c}{\textit{Click}} & \multicolumn{2}{c}{\textit{Termination}} & \multicolumn{2}{c}{\textit{Repeat}} \\
{} &\multicolumn{1}{c}{\textit{Strategy}}& \multicolumn{2}{c}{\textit{Score}} & \multicolumn{2}{c}{\textit{Agreement}} & \multicolumn{2}{c}{\textit{Agreement}} & \multicolumn{2}{c}{\textit{Agreement}} \\
Condition &  {} &      Median &    IQR &         Median &   IQR &        Median &   IQR &          Median &   IQR \\
\midrule
Choice Tutor    &    71.4\%  &             22.67 &  26.14 &           0.32 &  0.24 &          0.94 &  0.05 &             0.0 &  0.33 \\
No Tutor       &   16.1\%  &           11.46 &  19.68 &           0.12 &  0.25 &          0.93 &  0.07 &             0.0 &  0.00 \\
Dummy Tutor     & 43.1\%  &      14.11 &  23.34 &           0.25 &  0.30 &          0.92 &  0.09 &             0.0 &  0.00 \\
\bottomrule
\end{tabular}}
\caption{Median results and Inter Quartile Range (IQR) of participants' performance in the experiment. The \textit{goal planning} strategy measure is reported as the percentage of participants who learned the desired behavior.}
\label{table:res_median}
\end{table}

The results of the experiment are summarized in Table~\ref{table:res_median}. We start by comparing the unaided human performance (\textit{No Tutor} condition) to the performance of MGPO when evaluated on the same set of environments. MGPO's strategy achieves a median \textit{RR-score} of $22.70$ (interquartile range: $21.25$), compared to the lower \textit{RR-score} of $11.46$ achieved by participants in the \textit{No Tutor} condition. A Wilcoxon signed-rank test confirmed that this difference is significant ($T=102373$, $p<.001$), indicating that people's strategies for solving partially observable meta-level MDPs are suboptimal and can potentially be improved by teaching them MGPO's superior planning strategy. Comparing participants in the \textit{Choice tutor} to MGPO's strategy confirmed the effectiveness of the intelligent tutor. Participants in the \textit{Choice tutor} condition achieved a \textit{RR-score} of $22.67$ while MGPO achieved a \textit{RR-score} $22.41$ (interquartile range: $21.98$) on the same environments. Even though participants of the \textit{Choice Tutor} condition didn't learn to follow the demonstrated strategy exactly, training with the intelligent tutor taught them a similarly efficient planning strategy that numerically even slightly surpassed MGPO's strategy in the evaluation environments. Using a Wilcoxon signed-rank test, we did not find a significant difference between participants in the \textit{Choice tutor} condition and MGPO's strategy ($p=.453$). We now continue our analysis with a detailed comparison between the three conditions.

Participants' \textit{RR-scores} on the test trials differed significantly between the three conditions ($F(1.9,  259.58)=33.22$, $p<.001$). Post hoc ANOVA-type statistics showed that the performance of participants in the \textit{Choice Tutor} condition was significantly higher than the performance of participants in the \textit{No Tutor} ($F(1)=57.90$, $p<.001$) and participants in the \textit{Dummy Tutor} condition ($F(1)=50.52$, $p<.001$). There was no significant difference in the performance of the participants of the two control conditions ($F(1)=1.14$, $p<.29$).

The strategy discovered by our method followed a general goal-directed planning strategy
of first examining nodes the furthest away from the root node to select one or more potential goals (goal-setting). Planning backward from promising goals, it then inspects intermediate nodes, and lastly the nodes closest to the root node. 
Our analysis showed that significantly more participants in the \textit{Choice Tutor} condition learned the \textit{goal planning} component of MGPO's strategy than participants in the \textit{No Tutor} condition ($z=7.69$, $p<.001$) and participants in the \textit{Dummy Tutor} condition ($z=4.04$, $p<.001$). 

We investigated how closely participants followed MGPO's strategy using a measure called \textit{click agreement}.
We found significant differences in \textit{click agreement} between conditions using the ANOVA-type statistic with Box approximation \citep{box1954some} ($F(1.98, 283.42)=49.67$, $p<.001$). Post hoc ANOVA tests showed that participants in the \textit{Choice Tutor} condition had a significantly higher \textit{click agreement} than participants in the \textit{No Tutor} condition ($F(1)=107.42$, $p<.001$) and \textit{Dummy Tutor} condition ($F(1)=16.14$, $p<.001$). 

To evaluate how well participants learned when to stop planning, we compared when planning was terminated by participants versus MGPO's strategy. 
The ANOVA-type statistic with Box approximation \citep{box1954some} confirmed significant differences between conditions ($F(1.93, 267.49)=10.39$, $p<.001$). Post hoc ANOVA tests showed that participants in the \textit{Choice Tutor} condition achieved a significantly higher \textit{termination agreement} than participants in the \textit{No Tutor} condition ($F(1)=4.75$, $p=.029$) and \textit{Dummy Tutor} condition ($F(1)=23.63$, $p<.001$).

Lastly, we investigated if participants understood the value of investing additional computations by repeating their clicks as often as MGPO's strategy repeats computations. 
The ANOVA-type statistic with Box approximation \citep{box1954some} revealed significant differences in \textit{repeat agreement} between conditions ($F(1.96, 273.48)=15.41$, $p<.001$). Post hoc ANOVA-type statistics showed that participants in the \textit{Choice Tutor} condition achieved a significantly higher \textit{repeat agreement} than participants in the \textit{No Tutor} condition ($F(1)=22.45$, $p<.001$) and the \textit{Dummy Tutor} condition ($F(1)=25.33$, $p<.001$). 
Overall, participants struggled to learn when to repeat a computation, resulting in a median \textit{repeat agreement} of $0$. Differences between conditions are therefore only visible in the mean performance scores, with participants of the \textit{Choice tutor} condition achieving a mean repeat agreement of $0.18$, compared to $0.08$ for the \textit{No tutor} condition, and $0.09$ for the \textit{Dummy tutor} condition.

\subsection{Analysis of the discovered strategy} \label{sec:discovered_strategy}

\begin{figure}[ht]
  \centering
  \includegraphics[width=\textwidth]{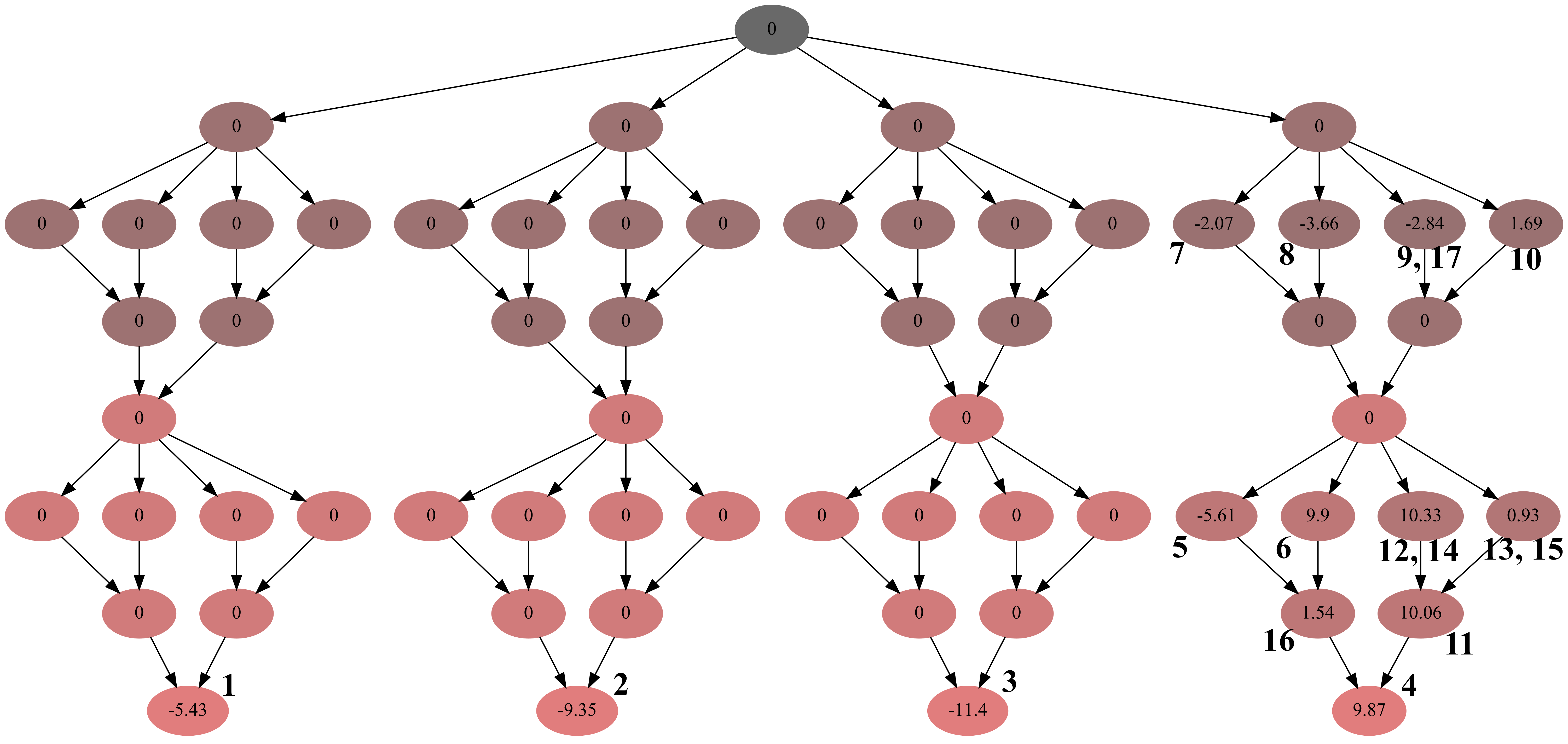}
  \caption{Example of the computed strategy in a specific environment instance. The numbers in the nodes show the expected reward of the nodes according to the current belief state. The node's color indicates the uncertainty, ranging from no uncertainty (gray) to high uncertainty (red). The numbers next to the nodes indicate in which order the nodes have been selected by the planning strategy discovered by our method.}
  \label{fig:env_strategy}
\end{figure}

To better understand the strategy used by our algorithm, we inspected the computations it performs in a sample environment (see Figure~\ref{fig:env_strategy} for the performed computations and Figure~\ref{fig:env_exp} for the reward distribution of the different nodes). In this environment, all paths to a final goal at the bottom of the graph have to pass through a distinct bottleneck in the middle of the graph. Therefore, each final goal is naturally associated with one specific subgoal. The planning strategy discovered by our algorithm follows a general pattern: 1) first selecting a final goal by inspecting the orange nodes in Figure~\ref{fig:env_exp}, 2) then selecting a path from the corresponding subgoal to the final goal by inspecting the green nodes in Figure~\ref{fig:env_exp}, and 3) planning how to reach the subgoal by inspecting the blue nodes in Figure~\ref{fig:env_exp}.  This order reflects that the variance of the reward distribution is highest for the final goals, second highest for the nodes between the subgoal and the final goal, and lowest for the nodes between the starting location and the subgoals. Inspecting the nodes between the subgoal and the final goal could, in principle, overturn the chosen final goal, but inspecting the nodes before the subgoal cannot.

In the first step, the strategy inspects potential final goals until a good goal has been identified. This can include repeating computations for one or more goals if they are sufficiently close in expected reward. In the second step, intermediate nodes leading to the selected goal are investigated until a sufficiently good subgoal has been identified. Here, again, nodes can be sampled multiple times if they show similar expected rewards. In the third and final step, the nodes close to the root node that lead to the selected goal are investigated. The algorithm then tries to improve upon the current best path by repeatedly sampling additional alternatives if their VOC is sufficiently high. 

Importantly, this example of MGPO's strategy is not a universal sequence of computations that MGPO performs in every environment, but a single example of how MGPO plans. Every selected computation other than the first depends on the randomly sampled observations and the (also randomly initialized) ground truth rewards of the observed nodes. If, for example, the first observation in the example environment (Figure~\ref{fig:env_strategy}) revealed a high enough value, MGPO might have skipped investigating the other goal nodes in the environment and immediately planned a path towards the first goal node.

\subsection{What did people learn from the tutor?}

Understanding the strategy discovered by MGPO allowed us to investigate which parts of the strategy participants in the experiment did and did not learn. To achieve this, we analyzed to which extent participants' click sequences exhibited the key three-step structure of the MGPO's strategy described in Section~\ref{sec:discovered_strategy}.  We already reported which percentage of participants follow the general structure of starting to plan by identifying a goal, then moving towards intermediate nodes, and ending with planning a path in immediate nodes in Section~\ref{exp_results} (see \textit{Goal Strategy}). This section provides additional detail investigating the errors participants made.

\paragraph{Setting a final goal}
We first evaluated how well participants learned the first step of the MGPO strategy (i.e., starting by selecting a final goal) using two measures. First, we compared the percentage of participants who started their planning by inspecting a goal node. In the \textit{Choice tutor} condition, $84\%$ of participants selected a goal node first, while only $55\%$ of the \textit{Dummy tutor} condition and $23\%$ of the \textit{No tutor} condition learned this part of the strategy. Especially in the \textit{No tutor} condition, participants failed to learn which nodes to prioritize and started with the least valuable node type, immediate nodes, in 64\% of test trials.

Second, we investigated which percentage of participants learned the full goal selection aspect of MGPO's strategy. To calculate this measure, we computed participants' \textit{click agreement} for consecutive goal node selections at the start of planning. In the \textit{Choice tutor} condition, participants reached a \textit{click agreement} of $57\%$, compared to $40\%$ for participants in the \textit{Dummy tutor} condition and $16\%$ for participants in the \textit{No tutor} condition. While the most common errors in the \textit{No tutor} and \textit{Dummy tutor} conditions were to not start with goal planning at all ($77\%$ and $45\%$ of test trials), in the \textit{Choice tutor} condition the most common error was overplanning which goal to pursue ($69\%$ of test trials), a less costly mistake since incurring an additional cost of computation results in a smaller decrease in \textit{RR-score} compared to not knowing which goal to pursue. 

\paragraph{Planning the path from the corresponding subgoal to the final goal}
The second step of the MGPO's strategy was to inspect intermediate nodes to plan how to get to the final goal from the corresponding subgoal (bottleneck). We, therefore, analyzed which percentage of participants learned to inspect an intermediate node after inspecting at least one final goal node but before inspecting any immediate nodes. In the \textit{Choice tutor} condition, $67\%$ of participants learned this aspect of the strategy, compared to $37\%$ in the \textit{Dummy tutor} condition and only $15\%$ in the \textit{No tutor} condition. The most common mistake made by participants in the \textit{Dummy tutor} ($33\%$ of test trials) and the \textit{No tutor} ($69\%$ of test trials) conditions was to plan intermediate nodes before planning which goal to pursue. The most common mistake of participants in the \textit{Choice tutor} condition was to select an intermediate node only after having already selected at least one immediate node ($19\%$ of test trials).

\paragraph{Planning how to reach the subgoal}
The third step of the MGPO strategy is to plan the path to the subgoal by inspecting immediate nodes after identifying a promising goal and intermediate nodes. The percentage of participants who, like the MGPO strategy, only selected an immediate node after already selecting at least one goal node and intermediate node was $69\%$ for the \textit{Choice tutor} condition, $41\%$ for the \textit{Dummy tutor} condition, and $17\%$ for the \textit{No tutor} condition. In both the \textit{Dummy tutor} condition ($31\%$ of test trials) and the \textit{No tutor} condition ($73\%$ of test trials), the most common mistake was to plan within immediate nodes before planning which goal to pursue. In the \textit{Choice tutor} condition, the most common mistake was failing to select an intermediate node before inspecting immediate nodes ($16\%$ of test trials). 

\bigskip
\noindent
Participants in the two control conditions made costly mistakes in their inspections of intermediate and immediate nodes because expending resources to plan a path to a potentially bad goal is highly inefficient. In comparison, the mistakes made by participants in the \textit{Choice tutor} condition were arguably less costly since the participant had already identified a worthwhile goal, and any nodes on the path to that goal have the potential of increasing the participant's \textit{RR-score}.

While the optimal strategy discovered by MGPO may seem relatively easy to understand and execute, it is important to note that the problem of discovering the strategy out of the space of all possible planning strategies is significantly harder. This is demonstrated by the fact that participants in the two control conditions failed to understand crucial components of the optimal strategy, such as planning which goal to pursue before investigating intermediate nodes. It may seem surprising that this aspect was not understood by a larger percentage of participants, but overall the suboptimality of participants' self-learned planning strategies is consistent with prior work on meta-planning \citep{Callaway2021Leveraging, skirzynski2021automatic, consul2021improving}. In fact, even on the significantly easier three-step planning problem with a similar structure used by \citet{Callaway2021Leveraging}, participants often failed to use backward planning strategies.

\section{Conclusion}

We have developed the first intelligent tutor for discovering and teaching resource-rational planning strategies for partially observable environments (MGPO). Our results show that the tutor improves people’s planning skills by teaching them the planning strategy discovered by MGPO. People trained by the intelligent tutor learned significantly better planning strategies compared to people who practiced by themselves or were trained with a similar tutor that doesn't rely on MGPO. We introduced three technical advances to make this possible: (1) extending the meta-level MDP model of optimal human planning strategies to partially observable environments, and (2) developing MGPO, a metareasoning algorithm that derives near-optimal strategies for human planning in partially observable environments, and (3) creating and evaluating a novel intelligent tutor that teaches the discovered planning strategies by approximating optimal metacognitive feedback. 


We believe intelligent tutoring systems based on strategy discovery methods are a promising technology for improving human decision-making in the real world. While our previous intelligent tutors \citep[i.e., ][]{Callaway2021Leveraging,consul2021improving,mehta2022leveraging} were effective at teaching humans better planning strategies, they were limited to highly simplified environments. Our new intelligent tutor based on MGPO extends this approach to partially observable environments. This is a crucial step towards improving human planning in the real world because most real-world situations are only partially observable. Although some prior work has successfully developed intelligent tutoring systems teaching metacognitive strategies in the real world, those systems have often relied on teaching hand-crafted heuristics to do so \citep{aleven2006toward, chi2010meta}. MGPO automates the process of discovering near-optimal metacognitive strategies that real people can execute in the real world by using a metacognitive model of the planning task.  
Additionally, intelligent tutoring systems that utilize strategy discovery methods can compute optimal metacognitive feedback based on the learner's errors and current information, which offers an effective way to automate teaching the discovered strategies as well.

Recent research suggests that people learn cognitive strategies partly 
through simple, model-free reinforcement learning mechanisms \citep{HeJainLieder2021,krueger2017enhancing}. If this is the case, then methods that have been developed to foster this simple form of learning, such as feedback \citep{Callaway2021Leveraging} and shaping \citep{skinner1953shaping,skinner1958reinforcement}, may be very appropriate for helping people learn how to make better decisions. Since our intelligent tutor was based on those principles, its effectiveness suggests that this might be the case. Regardless thereof, more cognitively sophisticated learning mechanisms are also crucial and should be supported as well.

MGPO was designed specifically for efficiently solving environments utilizing our model of the partially observable meta-level MDPs and therefore heavily utilizes the Gaussian nature of belief states to calculate the VOC. This can be seen as both a feature and a limitation. While the dependency on normally distributed environment models limits the types of environments MGPO can be applied to compared to prior methods built on categorical probability distributions (i.e. \citealp{callaway2017learning, consul2021improving}), it also enables MGPO to outperform existing methods in partially observable environments, for which no prior methods for solving meta-level MDPs are applicable. While it is possible to extend the method to more expressive reward distributions in future work, normal distributions are also widely present in nature \citep{lyon2014normal} and have been used extensively in relevant prior work \citep{dearden1998bayesian,guez2012efficient,mehta2022leveraging,consul2021improving,lieder2017automatic}.

MGPO further assumes that all meta-level actions are available at all stages of the planning process and can be executed independently of each other. Although we believe that this structure of planning tasks is applicable to many realistic settings, some aspects of real-world planning problems may require more complex models with sequentially dependent meta-level actions. To an extent, MGPO can already solve these cases with minor adjustments, such as interleaving meta- and object-level actions, or integrating multiple sequential planning operations into a single meta-level action. However, to fully address dependencies in meta-level actions, MGPO would need to be extended to plan multiple steps into the future, which will require additional optimizations due to the computational complexity. 

Additionally, our environment model is further constrained by the assumptions that (1) the variance of rewards is known, (2) the environment accurately models the effects of intermediate actions on the final decision, and (3) computations yield a noisy sample from the true reward distribution. These limitations, which are shared with prior work \citep{consul2021improving, Callaway2021Leveraging}, pose serious challenges when trying to apply strategy discovery to real-world scenarios. Besides estimating environment parameters from real-world data, one approach to dealing with potential uncertainty over environment parameters like the variance of rewards or the initial belief is to integrate prior work on robust strategy discovery \citep{mehta2022leveraging} into MGPO, which allows to optimize strategies over all possible instances of environment parameters using Bayesian inference. 

Our simplification that computations sample from the true reward distribution could in part be addressed by integrating a model of people's biases when performing mental simulations into the meta-level MDP. One such model is utility weighted sampling \citep{lieder2015utility}, which takes the common bias of overweighting extreme events into account. While these potential changes to reward sampling strategy provide a practical way to address common biases, ultimately, the usefulness of planning is inherently limited by the quality of people's mental simulations. Therefore, people's internal mental simulations being less accurate than the provided samples might make the strategies less beneficial for them. 

The flight planning problem used in our training experiment is an abstraction of the type of planning problems people face in the real world. Therefore, it does not yet qualify as a practical real-world educational application and does not directly match the kinds of problems students commonly learn to solve. Instead, our article introduces the general principles for the development of computer-based learning systems that automatically discover planning strategies in partially observable environments and teach them to people using feedback-based cognitive tutors. We believe that these principles also apply to the design of cognitive tutors for problem-solving in more realistic settings and the abstractions commonly used in standard curricula.

We hope that future work will apply the general principles introduced in this article to develop or improve intelligent tutoring systems for teaching the problem-solving skills students need to succeed in academic contexts, such as solving word problems in math and physics or planning how to write an essay,  and real-world settings, such as career planning, project management, and financial decision-making. 

The path from our online laboratory into schools might be long, but recent progress suggests that the principles introduced in this article can be applied to leverage AI to discover and teach cognitive strategies that teach people how to make better decisions in the real world \citep{heindrich2024leveraging, becker2022boosting}. Concretely, we have recently shown that these principles can be applied to discover and teach adaptive strategies for deciding between different mortgages \citep{becker2022boosting} and choosing between projects \citep{heindrich2024leveraging}. 

Three possible pathways toward real-world applications are (1) teaching transferable cognitive skills in abstract environments that capture the essential structure of specific real-world tasks, (2) teaching general cognitive skills that can be applied to various real-world contexts, and (3) teaching directly applicable strategies in an accurate meta-level MDP model of a real-world task. While there is evidence that learners can transfer strategies to different planning problems \citep{Callaway2021Leveraging}, modeling relevant real-world issues directly within the meta-level MDP framework is the more direct approach with less need for the transfer of strategies. Strategy discovery methods could then directly discover resource-rational planning strategies for these problems, and teaching them to people will have a large potential benefit to their lives. To make this possible, recent work has made strategy discovery methods more scalable \citep{consul2021improving}, more interpretable \citep{becker2022boosting, skirzynski2021automatic}, and more robust against errors and uncertainties in our models of the real-world \citep{mehta2022leveraging}.

Creating metareasoning models for real-world problems is a key difficulty in applying strategy discovery to the real world, as strategies discovered using inaccurate models might result in poor performance when applied to the real problem. In environments where relevant data is widely available, environment parameters can likely be directly estimated from data, with increasing confidence in the model's accuracy the more data is available. An adapted version of MGPO has been successfully applied to improve people's project selection decisions in follow-up work \citep{heindrich2024leveraging}, in which the environment parameters were estimated from real-world data. For environments without sufficient data availability, the model could be created by domain experts instead. As both methods of modeling real-world problems might introduce inaccuracies or biases into the model, combining MGPO with robust strategy discovery \citep{mehta2022leveraging} is an especially exciting direction for future work.

In summary, we have presented an intelligent tutor that leverages human-centered AI to improve human planning and decision-making in partially observable environments, and we have demonstrated its effectiveness through simulations and a training experiment. This research is an important step towards improving human decision-making in complex real-world tasks requiring farsighted planning under uncertainty. Our intelligent tutor makes it possible to teach highly complicated planning strategies to people. MGPO can discover strategies that are substantially better than the strategies people intuitively use. It can, therefore, be used to characterize the principles of good decision-making under uncertainty that we could previously only speculate about. 
MGPO is computationally efficient enough for interactive tutoring. Future work on intelligent cognitive tutors might make it feasible to cost-effectively boost the decision-making competence of millions of people. It might thereby become possible to improve crucial decisions without resorting to nudging \citep{hertwig2017nudging} and without interfering with people's freedom to make their own decisions \citep{thaler2021nudge}.

\subsection*{Statements and Declarations}

\paragraph{Acknowledgements}
The authors thank the International Max Planck Research School for Intelligent Systems (IMPRS-IS) for supporting Lovis Heindrich.

\paragraph{Funding}
This project was funded by grant number CyVy-RF-2019-02 from the Cyber Valley Research Fund.

\paragraph{Competing Interests}
The authors declare no competing interests.

\paragraph{Author contributions}
All authors designed the strategy discovery method and wrote the paper. Falk Lieder and Lovis Heindrich designed the experiment. Lovis Heindrich implemented the strategy discovery method, performed the experiment, and analyzed the data. 

\paragraph{Ethics approval}
The experiment was approved by the IEC of the University of T\"ubingen under IRB protocol number 667/2018BO2 (``Online-Experimente \"uber das Erlernen von Entscheidungsstrategien'') and was performed in accordance with the relevant guidelines and regulations.

\paragraph{Consent to participate}
All participants were over 18 years old, fluent in the English language, and gave informed consent to participate in the experiment.

\paragraph{Data availability} 
The code of the strategy discovery method as well as data and analysis for the simulation and training experiment are available online: \url{https://github.com/RationalityEnhancementGroup/MGPO-SSD}.

\bibliography{mybibliography}

\begin{appendices}

\section{Extended analysis of the simulation experiment}

In this section, we present the extended analysis of the simulation experiment (see Section~3.1). We found significant interaction effects between the used algorithm and the used environment ($F(15, 239952)=378.55$, $p<.001$), and between the used algorithm and the used cost value ($F(5, 239952)=6046.29$, $p<.001$). We did not find a three-way interaction effect between algorithm, cost, and environment ($p=.136$). We compared MGPO to the baselines for different cost values and environments using post-hoc tests with no p-value adjustment. 

MGPO significantly outperformed all baselines when the cost was $0.05$: the meta-greedy baseline ($p<.001$), and PO-UCT with 10 steps (($p<.001$), 100 steps ($p<.001$), 1000 steps ($p=.002$), or 5000 steps ($p=.032$). When the cost was $1$, MGPO performed significantly better than PO-UCT with 10 steps (($p<.001$), 100 steps ($p<.001$), 1000 steps ($p=.002$), and 5000 steps ($p=.007$). Although MGPO performed numerically better, there was no significant difference between MGPO and the meta-greedy baseline for a cost of $1$ ($p=.3537$).

We further compared MGPO to the baseline algorithms for the different environments, averaging across cost values. MGPO numerically outperformed the baseline algorithms across all environments. For the meta-greedy policy, this difference was significant for the 5 goal environment ($p=.014$), but not for the 2 goal ($p=.0.077$), 3 goal ($p=.0.065$), or 4 goal environments ($p=.0.066$). MGPO significantly outperformed PO-UCT with 10 or 100 steps across all environments (all $p<.001$). MGPO also significantly outperformed PO-UCT with 1000 steps in the 2 goal ($p=.0.019$), 3 goal ($p=.0.015$), and 5 goal ($p=.0.024$) environment, but not the 4 goal environment ($p=.0.082$). Compared to PO-UCT with 5000 steps, MGPO performed significantly better in the 2 goal environment ($p=.0.029$), but not the 3 goal ($p=.0.129$), 4 goal ($p=.0.217$), and 5 goal ($p=.0.055$) environment.

There was a numerical benefit of using MGPO over any baseline method across all the different configurations of environment and cost value.

\end{appendices}

\end{document}